\documentclass[lettersize,journal]{IEEEtran}
\usepackage{amsmath,amsfonts}
\usepackage{algorithmic}
\usepackage{algorithm}
\usepackage{array}
\usepackage[caption=false,font=normalsize,labelfont=sf,textfont=sf]{subfig}
\usepackage{textcomp}
\usepackage{stfloats}
\usepackage{url}
\usepackage{verbatim}
\usepackage{graphicx}
\usepackage{multirow}
\usepackage{color}
\usepackage{cite}
\usepackage{orcidlink}
\begin{document}

\title{An Efficient Aerial Image Detection with \\ Variable Receptive Fields}

\author{Wenbin Liu~\orcidlink{0009-0008-6626-0194}}
\thanks{The authors are with the School of Electronic Information and Electrical Engineering, Shanghai Jiao Tong University, Shanghai 201100, China (e-mail: bilibili@sjtu.edu.cn}

\maketitle

\begin{abstract}
Aerial object detection using unmanned aerial vehicles (UAVs) faces critical challenges including sub-10px targets, dense occlusions, and stringent computational constraints. Existing detectors struggle to balance accuracy and efficiency due to rigid receptive fields and redundant architectures. To address these limitations, we propose Variable Receptive Field DETR (VRF-DETR), a transformer-based detector incorporating three key components: 1) Multi-Scale Context Fusion (MSCF) module that dynamically recalibrates features through adaptive spatial attention and gated multi-scale fusion, 2) Gated Convolution (GConv) layer enabling parameter-efficient local-context modeling via depthwise separable operations and dynamic gating, and 3) Gated Multi-scale Fusion (GMCF) Bottleneck that hierarchically disentangles occluded objects through cascaded global-local interactions. Experiments on VisDrone2019 demonstrate VRF-DETR achieves 51.4\% mAP\textsubscript{50} and 31.8\% mAP\textsubscript{50:95} with only 13.5M parameters. This work establishes a new efficiency-accuracy Pareto frontier for UAV-based detection tasks. Our code is open-sourced and publicly available on \url{https://github.com/LiuWenbin-CV/VRF-DETR}.
\end{abstract}


\section{Introduction}
\IEEEPARstart{A}{erial} remote sensing target detection is crucial for applications in urban planning, environmental monitoring, and disaster response, where unmanned aerial vehicles (UAVs) capture high-resolution imagery for real-time object localization. However, UAV-based detection faces challenges such as small object sizes, occlusions due to dense object arrangements, and complex backgrounds that introduce noise. These difficulties necessitate the development of algorithms that strike a balance between accuracy, speed, and adaptability, especially for deployment on resource-constrained edge devices.

Traditional frameworks fall short in addressing UAV-specific complexities. Two-stage detectors like Faster R-CNN \cite{ren2016faster} rely on region proposal networks (RPNs) for high accuracy but suffer from excessive computational costs, rendering them impractical for real-time UAV deployment. Single-stage models such as YOLO \cite{redmon2016you} and SSD \cite{liu2016ssd} prioritize speed but depend on predefined anchor boxes and post-processing steps (e.g., non-maximum suppression), limiting flexibility in dynamic aerial scenes. Recent end-to-end detectors like DETR \cite{carion2020end} eliminate anchors and RPNs by leveraging transformer architectures for global context modeling. However, the computational complexity of DETR hinders its real-time performance.

Recent advancements in UAV object detection have introduced several enhanced YOLO and DETR variants, yet critical limitations persist in parameter efficiency, robustness, and fine-grained feature extraction. YOLO-DCTI \cite{min2023yolo} enhances small object detection in remote sensing through contextual transformer-based spatial-channel fusion and decoupled task-specific heads but increases model complexity after improved, limiting deployment on resource-constrained devices. UAV-DETR \cite{zhang2025uav} employs frequency-domain downsampling and semantic alignment to preserve high-frequency details but suffers from redundant parameters in multi-branch fusion structures, increasing computational overhead by 27\%. Drone-YOLO \cite{zhang2023drone} enhances detection heads with sandwich-fusion modules but lacks adaptive receptive field mechanisms, resulting in suboptimal performance for scale-varying targets in complex aerial scenes. HPS-DETR \cite{wang2025hps} reduces parameters through partial convolutions yet compromises fine-grained spatial information during channel selection, weakening discriminative power for occluded small objects. While RT-DETR \cite{zhao2024detrs} achieves real-time performance via hybrid encoders, its fixed-scale attention mechanisms fail to adaptively prioritize critical regions, leading low recall for sub-10px targets. These methods either escalate parameter counts through complex fusion strategies, employ rigid receptive fields insensitive to target scale variations, or sacrifice spatial details through aggressive feature compression, highlighting the need for lightweight architectures with adaptive perception and granular feature preservation.

To address these gaps, we propose Variable Receptive Field DETR (VRF-DETR), a transformer based detector tailored for UAV imagery. Our contributions are threefold:
\begin{enumerate}
\item{We propose the Multi-Scale Context Fusion (MSCF) module, which introduces adaptive receptive field selection and spatial attention to dynamically recalibrate multi-scale features, significantly improving small object detection accuracy while maintaining computational efficiency.}
\item{We design a Gated Convolution (GConv) module which integrates depthwise separable convolutions with dynamic gating mechanisms, achieving efficient spatial-context modeling with reduced parameters while preserving critical local patterns for real-time detection.}
\item{We develop the Gated Multi-scale Context Fusion (GMCF) Bottleneck, which cascades global attention and local gated operations in a hierarchical architecture to disentangle overlapping features from densely packed objects, particularly optimized for UAV imagery.}
\end{enumerate}

VRF-DETR outperforms state-of-the-art methods, achieving a 51.4\% mAP50 on VisDrone2019 with balanced speed and accuracy—a critical advancement for real-world UAV applications.

\begin{figure*}[!t]
\centering
\includegraphics[width=7in]{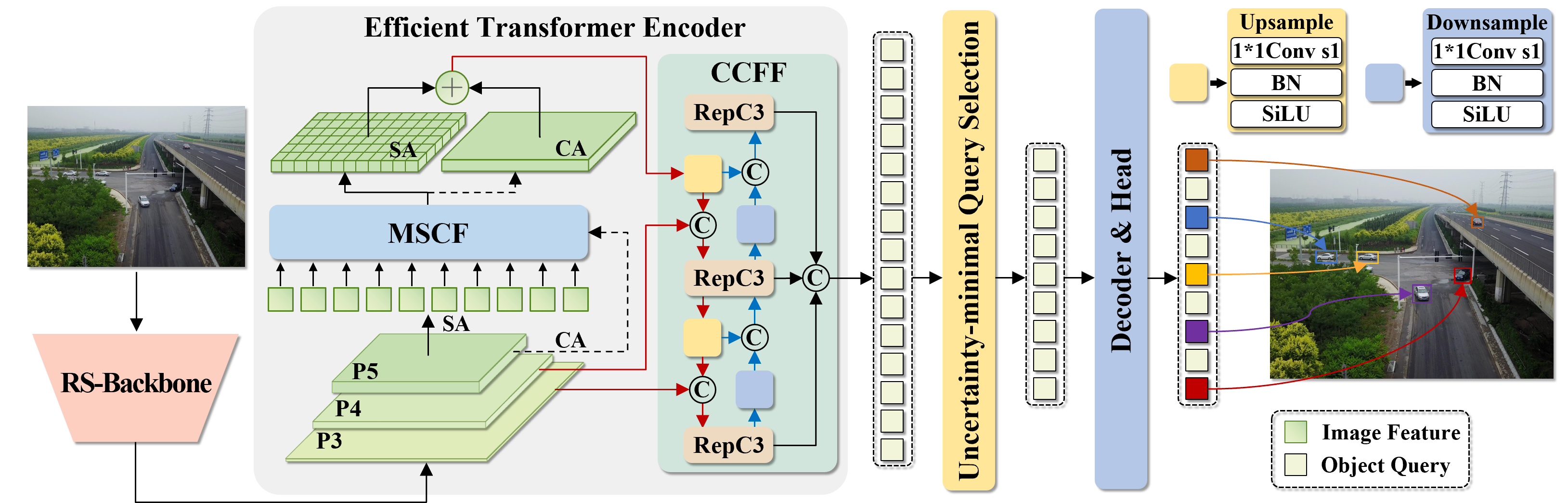}
\caption{Overall architecture of VRF-DETR. RepC3 is identical to that in RT-DETR. CA and SA representing channel and spatial attention, respectively.}
\label{fig_all}
\end{figure*}

\section{Methodology}
The proposed VRF-DETR architecture comprises three core components: a remote sensing backbone (RS-Backbone) with GMCF Bottleneck for hierarchical feature learning, a multiscale fusion encoder integrating adaptive spatial attention (MSCF), and lightweight GConv operators distributed network-wide for efficient spatial-context modeling, achieving accuracy-efficiency balance in aerial detection. The architecture of VRF-DETR is shown in Fig.~\ref{fig_all}.

\subsection{MSCF Module}
Wu et al. \cite{wu2023cmtfnet} propose a multi-scale multi-head self-attention (M2SA) module using dual branches to merge multi-scale context and channel attention. The multi-scale branch applies depthwise dilated convolutions with varied expansion rates (3, 5, 7), while the channel branch computes weights via feature compression-excitation \cite{hu2018squeeze}. Unlike traditional Transformer method, M2SA replaces raw inputs with multi-scale features as key-value pairs, lowering complexity to $O(nP)$. This structure boosts cross-scale feature fusion efficiently, enabling effective multi-scale modeling with low overhead. However, its fixed combination of multi-scale features without adaptive weighting may limit dynamic adaptation to objects of varying scales.

\begin{figure}[!t]
\centering
\includegraphics[width=3.5in]{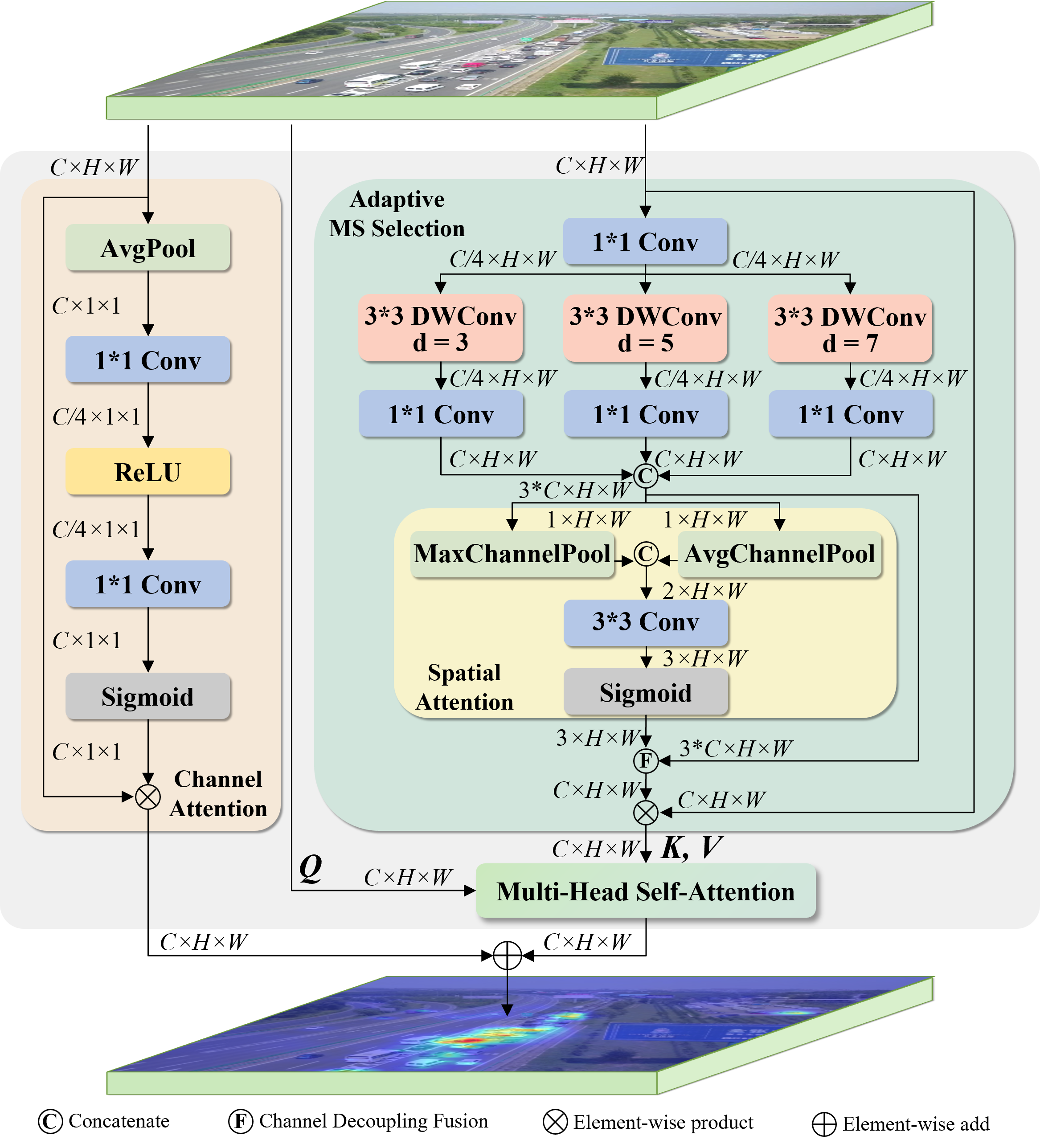}
\caption{The architecture of MSCF.}
\label{fig_MSCF}
\end{figure}

To address this limitation and enhance multi-scale representation in real-time detection architectures, we propose the Multi-Scale Context Fusion Attention Module (MSCF) as a lightweight replacement for the AIFI module in RT-DETR. Inspired by the optimization philosophy of the M2SA module, with a adaptive receptive field selection mechanism. As shown in Fig.~\ref{fig_MSCF}, MSCF uses a reorganized dual-branch structure. During multi-scale feature extraction, deep dilated convolutions are used to generate feature maps $F_i$ with varying receptive fields. Instead of the traditional direct summation method, we employ a three-step operation: feature concatenation, spatial selection, and weighted fusion. The features $F_i$ are concatenated along the channel dimension to form a multi-modal tensor, and a spatial attention mechanism generates a three-channel spatial selection mask $M$. This mask allows for differentiated weighting of the scale features $F_i$ via channel-wise decoupling. The final adaptive enhancement is achieved through gated element-wise multiplication. The process of the adaptive receptive field selection mechanism can be described as follows:
\begin{equation}
\label{eq:adaptive receptive field selection mechanism}
Y = \left( \sum_{i=1}^{N} \left( SA\left([F_1; \dots; F_N]\right)_i \odot F_i \right) \right) \odot X,
\end{equation}
where $\odot$ denotes Hadamard product and $SA$ refers to the spatial attention mechanism similarly in \cite{woo2018cbam}. $SA$ is defined as follows:
\begin{equation}
\label{eq:spatial attention mechanism}
SA = \sigma(\text{Conv}([P_{Avg}(F_{cat}); P_{Max}(F_{cat})])),
\end{equation}
where $\sigma$ denotes the Sigmoid activation function and $F_{cat}$ represents the concatenation of $[F_1; \dots; F_N]$.

By substituting the AIFI module in RT-DETR with MSCF, the network dynamically adjusts feature contributions based on target object scales and spatial distributions. The spatial attention mechanism not only improves focus on critical areas but also facilitates fine-grained interaction between multi-scale features. Additionally, the fusion process maintains resolution and channel consistency, ensuring lightweight characteristics. The proposed MSCF directly addresses the fixed-scale attention limitation of RT-DETR mentioned in Section I by implementing adaptive spatial weighting. This allows dynamic prioritization of critical regions containing sub-10px targets.

\subsection{GConv Module}
\begin{figure}[!t]
\centering
\includegraphics[width=2.5in]{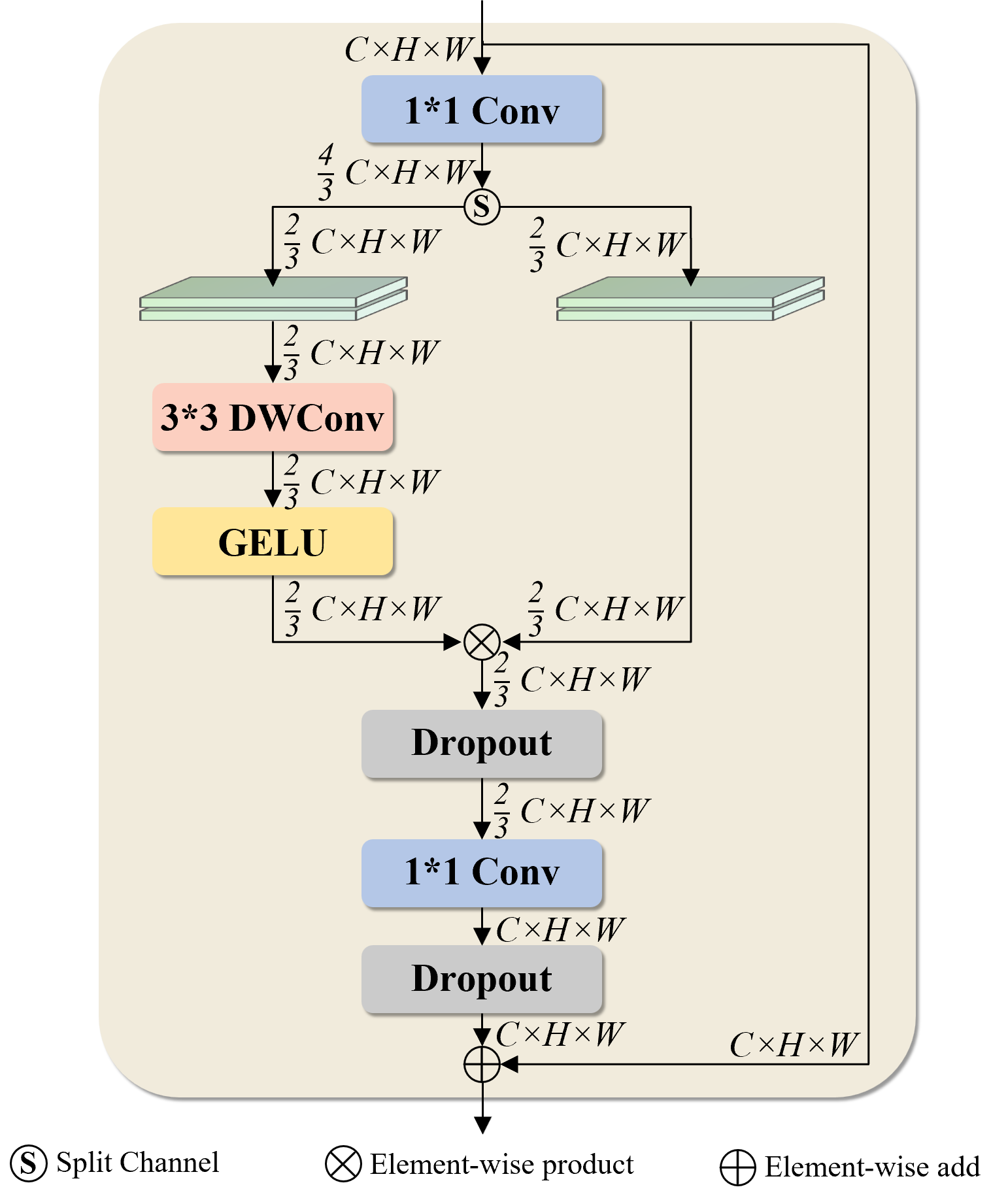}
\caption{The architecture of GConv Module.}
\label{fig_GConv}
\end{figure}

To balance feature extraction efficiency and parameter size in lightweight models, we propose a Gated Convolution Module (GConv), as shown in Fig.~\ref{fig_GConv}. GConv introduces an efficient gated feed-forward network augmented with spatial inductive bias, serving as an enhanced alternative to conventional feed-forward layers in vision architectures. The module comprises three key components: 1) A point-wise convolutional projection layer expanding input channels, 2) A parallelized depthwise convolution-based feature processor with gating control, and 3) A residual connection for identity mapping.

Mathematically, given input tensor $X \in \mathbb{R}^{C\times H\times W}$, the module first projects $X$ through a 1×1 convolution to double the channel dimension, subsequently splitting the output into two tensors $X^{'}, V \in \mathbb{R}^{C^{'}\times H\times W}$ via channel-wise chunking. The primary branch processes $X^{'}$ through a depthwise separable convolution with kernel size 3×3, followed by an activation function RELU. Spatial interactions are modulated through element-wise multiplication with the gating tensor $V$, implementing dynamic feature recalibration:
\begin{equation}
\label{eq:GConv}
\hat{X} = (\text{DWConv}(X^{'}) \cdot \sigma(1.702 \cdot \text{DWConv}(X^{'}))) \odot V,
\end{equation}
where $\odot$ denotes Hadamard product. This gating mechanism adaptively suppresses less informative spatial responses while preserving critical local patterns. The processed features then undergo channel restoration through a second 1×1 convolution and dropout regularization. Finally, a residual connection adds the original input $X$ to the transformed output.

Key innovations include: 1) $k \times k$ Depthwise convolution for spatial context aggregation with $O(C^{'}k^{2})$ parameters versus $O(C^{'2}k^{2})$ in standard convolutions, 2) GLU-inspired \cite{dauphin2017language} gating for input-dependent feature selection, and 3) Identity mapping to ease gradient flow. The hidden dimension $C^{'}$  is empirically set as $\frac{2}{3} C_{base}$ to balance computational overhead. This design effectively combines the global receptive field of 1×1 convolutions, spatial awareness from depthwise operations, and adaptive gating, making it particularly suitable for resource-constrained vision tasks. 

\begin{figure}[!t]
\centering
\includegraphics[width=2.5in]{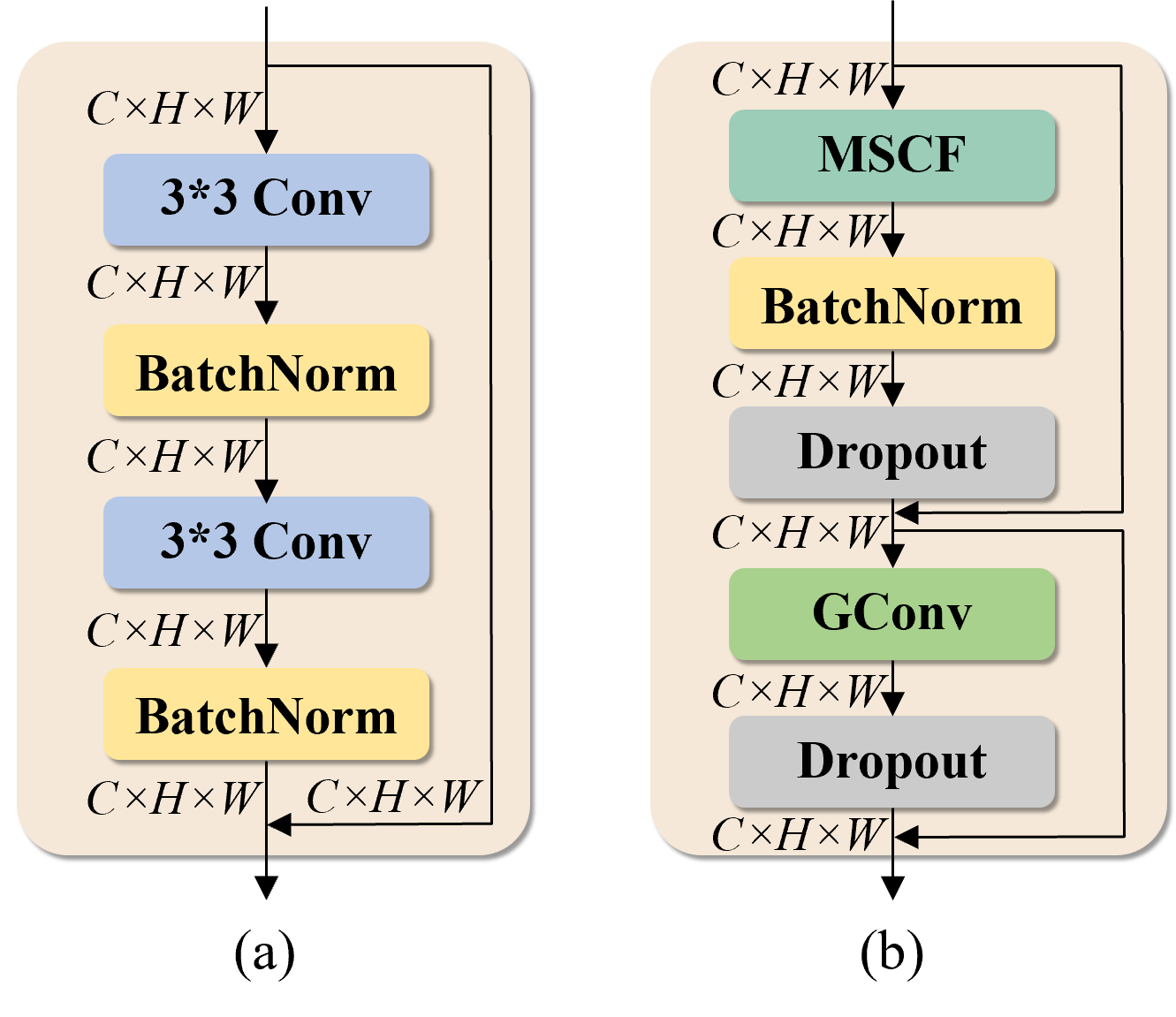}
\caption{Difference between C2f and GMCF Bottleneck. (a) C2f. (b) GMCF.}
\label{fig_GMCF}
\end{figure}

\subsection{GMCF Bottleneck}
In our approach, we replaced the original backbone of RT-DETR with RS-Backbone. The RS-Backbone architecture is similar to the backbone network of YOLOv8, but the multi-scale feature representation and fine-grained spatial feature capturing abilities of YOLOv8 are typically limited.  To strengthen multi-scale representation in RS-Backbone architectures, we propose the Gated Multi-scale Context Fusion (GMCF) Bottleneck to upgrade the C2f module, as depicted in Fig.~\ref{fig_GMCF}. This design integrates hierarchical attention mechanisms with gated convolutions, specifically optimized for capturing intricate patterns in aerial imagery. The original C2f module employs homogeneous feature processing, limiting its multi-scale modeling capacity. Our GMCF Bottleneck introduces a cascade structure: 1) MSCF establishes inter-region correlations, 2) Batch normalization and dropout layers regularize features, followed by 3) GConv for spatial-adaptive refinement. Dual shortcut connections flanking the cascade enable stable gradient flow, while intermediate dropout layers prevent co-adaptation. This sequential global-to-local processing hierarchy effectively disentangles overlapping features from densely packed small objects in aerial imagery. By incorporating the GMCF Bottleneck, we improved multi-scale feature representation and more effective spatial feature extraction, resulting in a more robust RS-Backbone.

\begin{table*}[!ht]
\centering
\caption{Comparison between VRF-DETR and State of the Art on the VisDrone-2019-DET val dataset. For each metric, the best result is highlighted in red bold, and the second-best result is highlighted in blue bold}
\label{tab:VisDrone}
\begin{tabular}{lcccccc}
\hline
\multirow{3}{*}{Model} & \multirow{3}{*}{Publication} & \multirow{3}{*}{Imgsize} & \multicolumn{2}{c}{VisDroneDET 2019} & \multicolumn{2}{c}{Model complexity} \\
& & & $\text{mAP}_{50}$ & $ \text{mAP}_{50-95}$ & Params & FLOPs \\
& & & (\%)$\uparrow$ & (\%)$\uparrow$ & (M)$\downarrow$ & (G)$\downarrow$ \\ 
\hline
\textbf{\textit{Two-stage methods}} & & & & & & \\
Faster R-CNN \cite{ren2016faster} & NeurIPS 2015 & 640$\times$640 & 31.0 & 18.2 & 41.2 & 127.0\\
ARFP \cite{wang2022arfp} & Appl Intell & 1333$\times$800 & 33.9 & 20.4 & 42.7 & 193.8\\
EMA \cite{ouyang2023efficient} & ICASSP 2023 & 640$\times$640 & 49.7 & 30.4 & 91.2 & -\\
SDP \cite{ma2023scale} & TGRS & 800$\times$800 & \textcolor{red}{\textbf{52.5}} & 30.2 & 96.7 & - \\
\hline
\textbf{\textit{One-stage methods}} & & & & & & \\
YOLOv12-M \cite{tian2025yolov12} & arXiv 2025 & 640$\times$640 & 43.1 & 26.3 & 20.2 & 67.5\\
HIC-YOLOv5 \cite{tang2024hic} & ICRA 2024 & 640$\times$640 & 44.3 & 26.0 & \textcolor{red}{\textbf{10.5}} & \textcolor{red}{\textbf{31.2}}\\
MSFE-YOLO-L \cite{qi2024msfe} & GRSL & 640$\times$640 & 46.8 & 29.0 & 41.6 & 160.2\\
YOLO-DCTI \cite{min2023yolo} & Remote Sensing & 1024$\times$1024 & 49.8 & 27.4 & 37.7 & -\\
Drone-YOLO-L \cite{zhang2023drone} & Drones & 640$\times$640 & 51.3 & 31.1 & 76.2 & -\\
\hline
\textbf{\textit{End-to-end methods}} & & & & & & \\
DETR \cite{carion2020end} & ECCV 2020 & 1333$\times$750 & 40.1 & 24.1 & 40.0 & 187.1\\
Deformable DETR \cite{zhu2020deformable} & ICLR 2020 & 1333$\times$800 & 43.1 & 27.1 & 40.2 & 172.5\\
RT-DETR-R18 \cite{zhao2024detrs} & CVPR 2024 & 640$\times$640 & 47.2 & 28.7 & 20.2 & 57.0\\
RT-DETR-R50 \cite{zhao2024detrs} & CVPR 2024 & 640$\times$640 & 50.7 & 30.9 & 41.8 & 133.2\\
UAV-DETR-R50 \cite{zhang2025uav} & arXiv 2025 & 640$\times$640 & 51.1 & \textcolor{blue}{\textbf{31.5}} & 42.0 & 170\\
\hline
\textbf{\textit{Ours}} & & & & & & \\
VRF-DETR & - & 640$\times$640 & \textcolor{blue}{\textbf{51.4}} & \textcolor{red}{\textbf{31.8}} & \textcolor{blue}{\textbf{13.5}} & \textcolor{blue}{\textbf{44.3}}\\
\hline
\end{tabular}
\end{table*}

\begin{table*}[!ht]
\centering
\caption{The comparison results of the model before and after improvement on the DOTA v1.0 val dataset. \\ The best result is highlighted in bold}
\label{tab:DOTA}
\begin{tabular}{lcccccc}
\hline
{Model} & $\text{mAP}_\text{S}(\%)\uparrow$ & $\text{mAP}_\text{M}(\%)\uparrow$ & $\text{mAP}_\text{L}(\%)\uparrow$ & $\text{mAP}_{50}(\%)\uparrow$ & $\text{mAP}_{50-95}(\%)\uparrow$ \\ \hline
RT-DETR-R50 \cite{zhao2024detrs} & 21.1 & 44.0 & 51.6 & 65.1 & 41.7 \\
VRF-DETR & \textbf{23.2} & \textbf{47.3} & \textbf{53.8} & \textbf{68.6} & \textbf{43.8}\\ \hline
\end{tabular}
\end{table*}

\begin{table}[!ht]
\centering
\caption{Hyperparameter Configuration}
\label{tab:hyperparameters}
\setlength{\tabcolsep}{4pt}
\begin{tabular}{>{\centering\arraybackslash}p{1.6in}>{\centering\arraybackslash}p{0.8in}} 
\hline
\textbf{Hyperparameter} & \textbf{Value} \\ \hline
Input size & 640$\times$640 \\ 
Batch size & 8 \\ 
Training epochs & 300 \\ 
Optimizer & AdamW \\ 
Initial learning rate & 0.0001 \\ 
Learning rate factor & 0.01 \\ 
Momentum & 0.9 \\
Warmup steps & 2000 \\ \hline
\end{tabular}
\end{table}

\section{Experiments}
\subsection{Datasets}
To validate VRF-DETR's effectiveness, we conducted experiments on VisDrone 2019 and DOTA v1.0 datasets. VisDrone 2019 contains UAV-captured images with challenges including uneven illumination and object occlusion, covering 10 categories such as pedestrians, cars, and buses. The official data splits were adopted, with evaluation conducted on the validation set. DOTA v1.0 features high-altitude aerial images with scale variations and class imbalance, containing 15 categories including planes, ships, and vehicles. Following standard protocols, experiments were evaluated on the validation set due to server unavailability.

\subsection{Experimental Configuration}
The experimental platform consists of a Windows 11 system with Python 3.8 and PyTorch framework. Hardware specifications include:
\begin{itemize}
\item CPU: 13th Gen Intel® Core™ i7-13700K @ 3.40GHz
\item GPU: NVIDIA GeForce RTX 4090 with 24GB GDDR6X memory
\end{itemize}

Detailed hyperparameter settings during training are presented in Table~\ref{tab:hyperparameters}. All experiments employ identical initialization strategies and data augmentation protocols to ensure fair comparison.

\subsection{Comparison Experiments}
\begin{figure}[!t]
\centering
\includegraphics[width=3.5in]{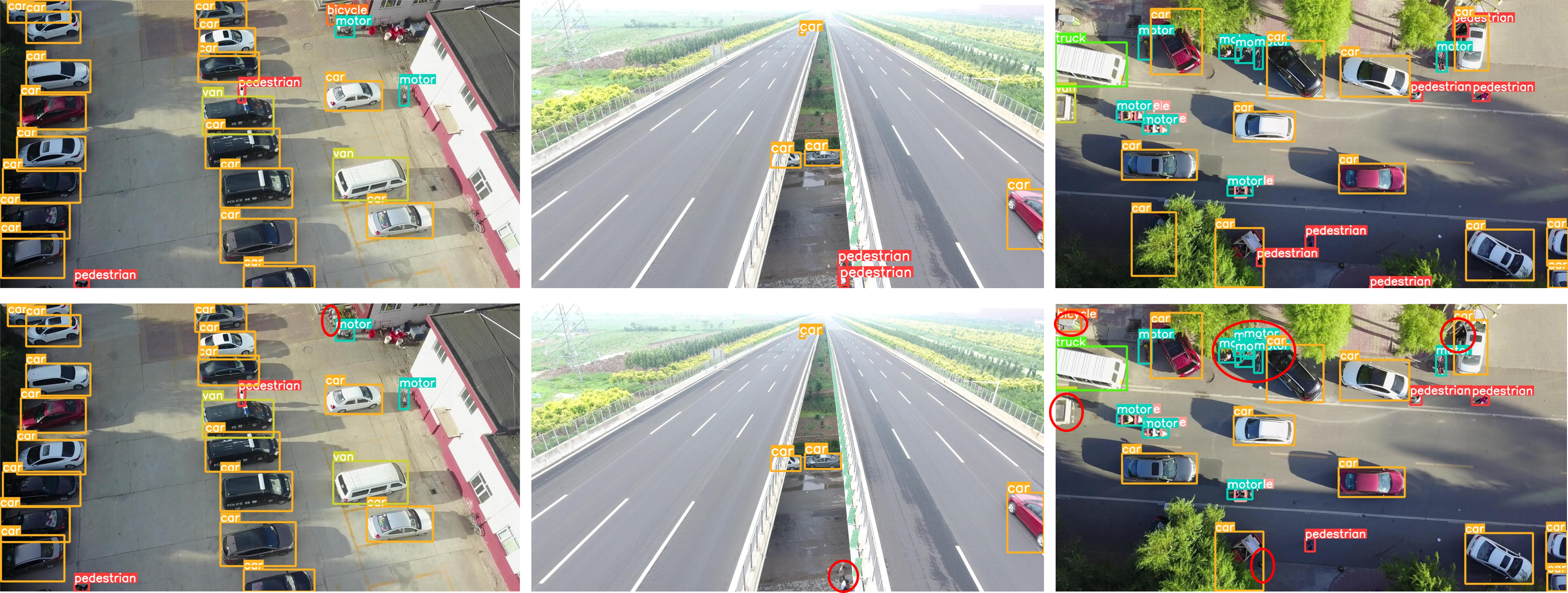}
\caption{Comparison of detection performance between VRF-DETR(the first row) and RT-DETR(the second row). Red boxes highlight incorrect detections.}
\label{fig_Results}
\end{figure}

As shown in Tables~\ref{tab:VisDrone} and~\ref{tab:DOTA}, our VRF-DETR achieves state-of-the-art $\text{mAP}_{50-95}$ on VisDrone2019 and outperforms RT-DETR on DOTA v1.0. Meanwhile, on the VisDrone2019 dataset, the $\text{mAP}_{50}$ of our model approaches that of the larger-input two-stage SOTA model SDP, while maintaining only 13.9\% of its parameter size. The framework demonstrates superior capability in resolving class ambiguity (e.g., car-van differentiation) and maintaining 60-80\% occlusion robustness, attributed to its adaptive receptive fields that overcome fixed-scale limitations. Notably, it achieves 3.2\% mAP improvement for medium-sized DOTA objects compared to baseline methods. The detection results comparisons in Fig.~\ref{fig_Results} further demonstrate our improved performance for small objects and severely occluded objects.

\begin{table}[!ht]
\centering
\caption{The ablation results based on the RT-DETR baseline were \\ conducted on the VISDRONE2019 validation dataset. \\ The best results are highlighted in bold.}
\label{tab:ablation}
\setlength{\tabcolsep}{4pt}
\begin{tabular}{
    >{\centering\arraybackslash}p{0.3in}
    >{\centering\arraybackslash}p{0.6in}
    >{\centering\arraybackslash}p{0.5in}
    >{\centering\arraybackslash}p{0.3in}
    >{\centering\arraybackslash}p{0.4in}
    >{\centering\arraybackslash}p{0.4in}
}
\hline
\multirow{2}{*}{MSCF} & 
\multicolumn{2}{c}{GConv} & 
\multirow{2}{*}{GMCF} & 
{$\text{mAP}_{50}$} & 
{$\text{mAP}_{50-95}$} \\
& in Backbone & in RepC3 & & $(\%)\uparrow$ & $(\%)\uparrow$ \\ 
\hline 
- & - & - & - & 47.2 & 28.7 \\
\checkmark & - & - & - & 49.0 & 29.8 \\
- & \checkmark & - & - & 47.4 & 28.7 \\
- & - & \checkmark & - & 47.0 & 28.6 \\
- & - & - & \checkmark & 49.5 & 30.2 \\
\checkmark & - & - & \checkmark & 50.1 & 30.5 \\
\checkmark & \checkmark & - & \checkmark & \textbf{51.3} & \textbf{31.7} \\
\checkmark & - & \checkmark & \checkmark & 50.6 & 30.8 \\ 
\checkmark & \checkmark & \checkmark & \checkmark & 51.1 & 31.5 \\
\hline
\end{tabular}
\end{table}

\subsection{Ablation Experiments}
As shown in Tables~\ref{tab:ablation}, the ablation study on VisDrone2019 validates our modules' efficacy. Baseline RT-DETR achieves 47.2\% mAP50. MSCF alone improves mAP50 by 1.8\%, while GMCF yields 49.5\%. Combining MSCF and GMCF reaches 50.1\%. Optimal performance emerges when integrating GConv in the backbone with both modules, surpassing the baseline by 4.1\% and 3.0\%. Full-model integration attains 51.1\% mAP50. Results confirm multi-scale fusion and gated convolutions strategically enhance drone detection.

\section{Conclusion}
This paper proposes VRF-DETR, a transformer-based framework addressing UAV detection challenges through three novel components. The MSCF module leverages adaptive spatial attention and gated fusion to dynamically recalibrate multi-scale features, resolving scale ambiguity in dense aerial scenes. The GConv module integrates depthwise separable convolutions with dynamic gating, enabling efficient local-context modeling while preserving discriminative spatial patterns critical for sub-10px targets. The GMCF Bottleneck cascades global attention and gated convolutions in a hierarchical architecture, disentangling overlapping features through sequential global-local interaction. Collectively, these modules establish a lightweight paradigm that adaptively prioritizes critical regions, suppresses background interference, and maintains fine-grained spatial details without heuristic anchors or NMS. The architecture’s parameter-efficient design and adaptive receptive fields offer a principled solution for UAV detection under computational constraints, balancing accuracy and real-time efficiency.

\bibliographystyle{IEEEtran}
\bibliography{References}

\end{document}